\documentclass[letterpaper]{article}

\usepackage[preprint]{aaai2027}
\usepackage[hyphens]{url}
\usepackage{graphicx}
\usepackage{natbib}
\usepackage{caption}
\usepackage{amsmath,amssymb}
\usepackage{booktabs}
\usepackage{multirow}
\usepackage{xcolor}
\usepackage{array}

\newcommand{\appDatasetProtocol}{C.1}
\newcommand{\appControlledAblations}{C.2}

\newcommand{\appOrderRobustness}{C.4}
\newcommand{\appCrossDataset}{C.5}
\newcommand{\appVisualAnalysis}{C.6}

\title{DARAD: Dual Adapters and Ranking-Aware Distillation for Continual Remote Sensing Image-Text Retrieval}
\author{
Xi Chen\textsuperscript{\rm 1},
Xu Chen\textsuperscript{\rm 1},
Xiangyang Jia\textsuperscript{\rm 1},
Wei Wang\textsuperscript{\rm 2},
Xu Zhang\textsuperscript{\rm 1},
Zhenyuan Sun\textsuperscript{\rm 1}
}
\affiliations{
\textsuperscript{\rm 1}School of Computer Science, Wuhan University, Wuhan 430072, China\\
\textsuperscript{\rm 2}Beijing Institute for General Artificial Intelligence (BIGAI)\\
\url{2024102110094@whu.edu.cn}, \url{xuchen@whu.edu.cn}, \url{jxy@whu.edu.cn},\\
\url{wangwei@bigai.ai}, \url{zhangx0802@whu.edu.cn}, \url{zhenyuansun@whu.edu.cn}
}

\begin{document}
\maketitle

\begin{abstract}
	With the rapid growth of Earth observation technologies, remote sensing archives are rapidly expanding, making remote sensing image-text retrieval (RS-ITR) increasingly important.
	However, continual RS-ITR remains challenging because scale variation and distribution shifts in RS aggravate cross-modal alignment space distortion, making it difficult for existing continual learning (CL) methods to support reliable continual retrieval.
	To address this challenge, we propose DARAD, a dual-adapter and ranking-aware distillation framework that preserves the historical cross-modal ranking structure while learning new visual and textual concepts from evolving archives.
	Specifically, the visual branch introduces a spatial fusion adapter, which integrates coarse regional cues and fine-grained patch cues to accommodate RS scale variation while anchoring visual updates to the pretrained alignment space.
	The textual branch employs multi-expert semantic routing, which separates shared textual semantics from semantically specialized residuals to absorb newly emerging descriptions while constraining global text embedding drift.
	Furthermore, bidirectional ranking distillation uses a frozen teacher model and historical anchors to preserve the historical cross-modal ranking structure, thereby mitigating alignment space distortion across continual stages.
	Experiments under a multi-stage continual retrieval protocol show that DARAD achieves superior performance over existing CL methods, improving adaptation to newly arrived data while maintaining effectiveness on historical data.
\end{abstract}
\section{Introduction}
	\label{sec:introduction}
	With the continuous advancement of Earth observation technologies, remote sensing (RS) image-text archives are rapidly expanding, making RS image-text retrieval (RS-ITR) increasingly important for cross-modal access to large-scale RS data~\cite{yuan2022amfmn,muhammad2025foundation}.
	Recent RS-ITR studies have progressed from both data and model perspectives: large-scale RS image-text datasets, including RS5M, SkyScript, and RST2I-110K, promote cross-modal representation learning~\cite{zhang2024rs5m,khan2024skyscript,zhang2026any2rsi}, while CLIP-like and RS-oriented vision-language models (VLMs) further improve RS-ITR and broader RS vision-language understanding through aligned image-text representations~\cite{liu2024remoteclip,liu2025cclip}.
	However, most existing RS-ITR methods assume a static data setting with fixed training data, semantic categories, and retrieval galleries; when newly arrived data are introduced, direct fine-tuning often improves current-data adaptation but degrades historical retrieval performance due to visual forgetting and text embedding drift.
	
	\begin{figure}[t]
	\centering
	\includegraphics[width=\linewidth]{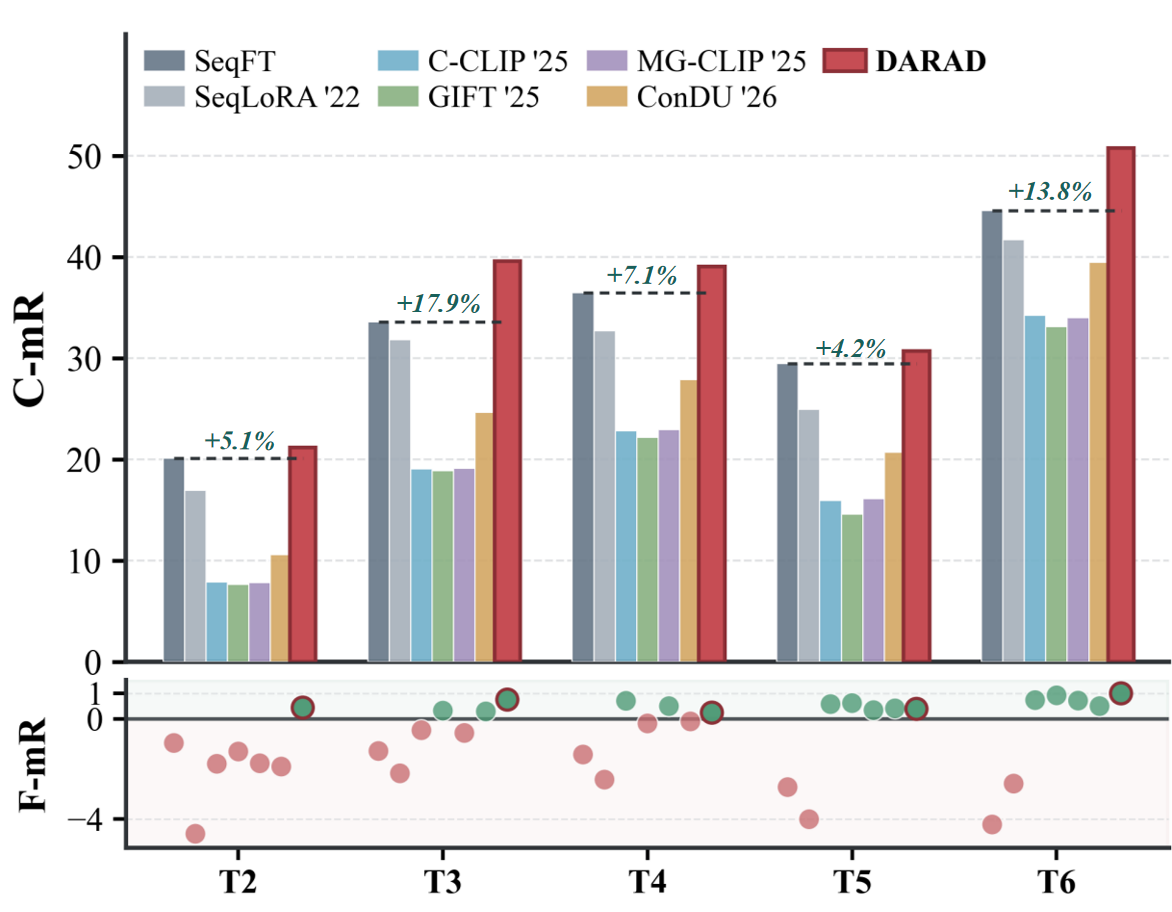}
	\caption{Current-stage mR (C-mR) over stages, with green annotations showing DARAD's gains over the baseline. The lower strip shows signed historical-change mR (F-mR): green/red markers indicate non-negative/negative change.}
	\label{fig:intro_continual_comparison}
	\end{figure}
	
	Although continual learning (CL) has been extensively studied for classification, detection, and recently VLMs, directly transferring existing CL strategies to semantic-stage continual RS-ITR remains insufficient~\cite{liu2025cclip,gao2026condu,wang2026vlmclsurvey}.
	This is mainly caused by three coupled challenges in continual RS-ITR.
	(1) Scale-induced visual drift. RS imagery contains objects and scenes with drastically different geographic scales, forcing the visual encoder to readjust spatial receptive fields when learning new tasks and thereby disturbing historical visual representations.
	(2) Asymmetric cross-modal drift. Textual semantics often undergo abrupt shifts due to terminology changes (e.g., from ``flooded area'' to ``commercial zone''), while visual feature distributions evolve more smoothly, invalidating previously established cross-modal alignments.
	(3) Ranking-sensitive retrieval structure. RS-ITR relies on relative rankings in the cross-modal embedding space, which conventional feature-level distillation designed for classification or detection does not explicitly preserve.
	
	To address these challenges, we propose DARAD, a dual-adapter and ranking-aware distillation framework for semantic-stage continual RS-ITR over evolving RS archives.
	In the visual branch, a spatial fusion adapter (SFA) integrates coarse regional cues and fine-grained patch cues to accommodate RS scale variation. 
	In the textual branch, multi-expert semantic routing (MSR) decomposes textual adaptation into a shared low-rank semantic basis and routed semantically specialized residual experts, enabling the model to absorb newly emerging descriptions while constraining global text embedding drift.
	In addition, bidirectional ranking distillation (BRD) uses a frozen teacher and compact historical anchors to align T2I, I2T, and anchor-internal relation matrices, explicitly preserving the historical cross-modal ranking structure against retrieval degradation caused by feature drift.
	As shown in Fig.~\ref{fig:intro_continual_comparison}, DARAD consistently improves current-stage retrieval over the strongest CL baseline while maintaining non-negative historical change across the continual stages.
	The main contributions of this work are as follows:
	
	(1) We propose DARAD, a dual-adapter method for semantic-stage continual RS-ITR, where SFA performs scale-aware visual adaptation and MSR performs semantically routed adaptation to mitigate visual forgetting and text embedding drift.
	
	(2) We introduce a BRD loss to preserve the historical cross-modal ranking structure during the dynamic expansion of the cross-modal alignment space.
	
	(3) We establish a multi-stage continual retrieval protocol for RS-ITR and conduct extensive experiments to evaluate new-data adaptation and historical retrieval performance retention over evolving RS image-text archives.
	
	\section{Related Work}
	\label{sec:related_work}
	\subsection{Remote Sensing Image-Text Retrieval}
	\label{subsec:rs_itr}
	
	RS-ITR aims to retrieve relevant RS images or texts across modalities by learning aligned image representations and text representations. Early studies were mainly developed on manually annotated benchmarks such as RSICD, UCM-Captions, and RSITMD~\cite{lu2018rsicd,yang2010ucmerced,qu2016ucm,yuan2022amfmn}, following long-standing RS image understanding and captioning research~\cite{zhang2024captionreview}. A series of task-specific matching methods were proposed to improve cross-modal alignment between RS images and texts~\cite{yuan2022amfmn,yuan2022galr,yuan2023pir}. These methods explored global-local matching, multi-scale attention, fine-grained visual-semantic alignment, and knowledge-aware representation learning to handle the complex spatial layouts, multi-scale targets, redundant regions, and semantic gaps in RS scenes~\cite{chen2024ebaker,li2024implicitexplicit,chen2024globallocal,wang2025finegrained,zhang2025contextaware,zheng2025fssn}. Recent variants further address representation discrepancy, modality gap, hashing efficiency, visual grounding, false-negative mitigation, and hierarchical or collaborative semantic alignment in RS cross-modal retrieval~\cite{zhao2025rdb,zavras2025modalitygap,li2024hashing,li2025visualgrounding,zheng2026trisim,zhou2026hierarchical,li2026explicitimplicit}. More recently, large-scale RS image-text datasets, such as RS5M, SkyScript, and Vigen500k, together with RS-oriented VLMs such as RemoteCLIP and GeoRSCLIP, have further promoted RS-ITR from task-specific supervised matching toward VLM-based representation learning~\cite{liu2024remoteclip,zhang2024rs5m,khan2024skyscript,ma2024vigen500k}. Several recent methods also introduce prompt learning, prior knowledge, keyword reasoning, coarse-to-fine retrieval, or entity-level semantic modeling to enhance fine-grained cross-modal matching~\cite{cheng2024air,li2025promptgranularity,li2025strongweak,zhang2025uncertainty,wang2025progressive,zhou2024cftir,wang2026iebaker}. Meanwhile, RS VLMs and multimodal foundation models have broadened RS vision-language learning beyond retrieval to grounded dialogue, instruction following, and multi-sensor understanding~\cite{kuckreja2024geochat,zhang2024earthgpt,zhan2024skyeyegpt,bazi2024rsllava,wang2024cogvlm,wang2025ddfav}. Despite these advances, most existing RS-ITR methods are still developed and evaluated under a static data setting, where the training data, semantic distribution, and retrieval gallery are assumed to be fixed. They do not explicitly address evolving RS image-text archives, where newly arrived data are continuously collected and direct fine-tuning may cause visual forgetting and text embedding drift, thereby degrading retrieval performance on historical data.
	
	\subsection{Continual Learning for Retrieval} 
	\label{subsec:continual_cross_modal}
	
	CL aims to enable models to learn from sequentially arriving data while alleviating catastrophic forgetting of previously acquired knowledge. Existing methods are commonly developed through replay, regularization, knowledge distillation, and architecture adaptation, and have been widely studied in classification, detection, and representation learning tasks. Beyond unimodal scenarios, CL has also been introduced into cross-modal retrieval and vision-language learning. Early continual cross-modal retrieval studies investigate how new tasks affect embedding spaces and cross-modal alignment, and further analyze the training, indexing, and querying stages of retrieval systems~\cite{ni2023modx,zheng2023zeroshotdegradation}. More recently, prompt learning, visual prompt tuning, multimodal prompt learning, and adapter-based tuning have provided lightweight ways to adapt CLIP-like VLMs~\cite{zhou2022coop,jia2022vpt,khattak2023maple,hu2022lora}, while continual VLM methods focus on adapting pretrained image and text encoders to new tasks or domains without severely degrading previously learned vision-language knowledge~\cite{yu2024moeadapters,liu2025cclip,wu2025gift,huang2025mgclip,gao2026condu,wang2026vlmclsurvey}. These studies provide important foundations for continual cross-modal representation learning. However, most of them are designed for general image-text data or generic VLM scenarios. Compared with natural images, RS images usually exhibit more complex spatial layouts and multi-scale region compositions, making continual RS-ITR require more careful control of visual forgetting and text embedding drift over evolving RS image-text archives.
	
	\section{Methodology}
	\label{sec:methodology}
	
	
	\subsection{Overview of DARAD}
	\label{subsec:overview}
	
\begin{figure*}[t]
		\centering
		\includegraphics[width=\textwidth]{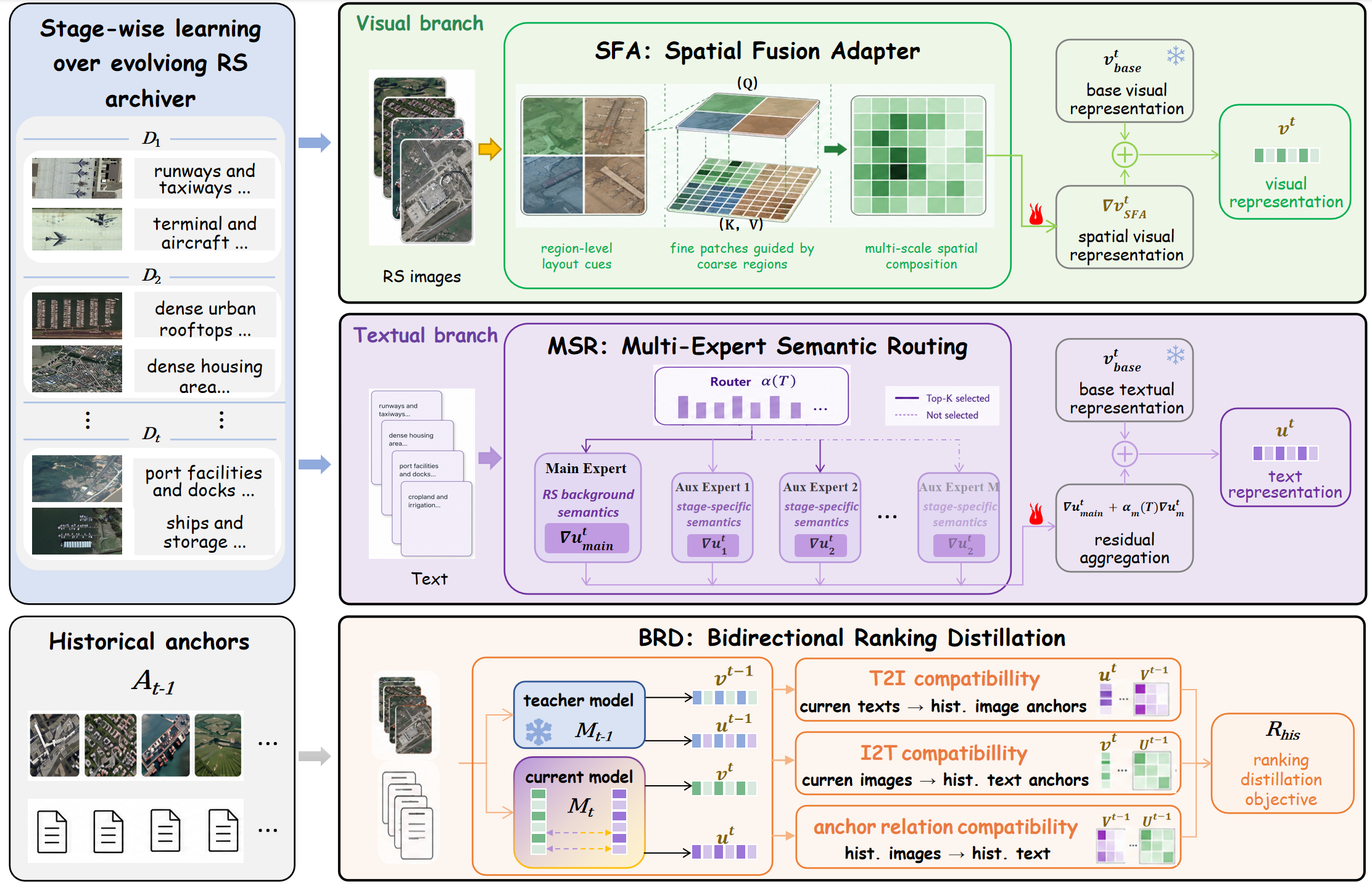}
		\caption{Overview of DARAD. SFA and MSR provide scale-aware visual and semantically routed textual adaptation, respectively, while BRD preserves the historical cross-modal ranking structure.}
		\label{fig:overview}
	\end{figure*}

	As illustrated in Fig.~\ref{fig:overview}, DARAD combines two complementary adapters with ranking-aware distillation for semantic-stage continual RS-ITR.
	SFA integrates coarse regional context and fine-grained patch cues for scale-aware visual adaptation, while MSR routes shared and semantically specialized residuals for controlled textual adaptation.
	BRD further distills bidirectional relation matrices from a frozen teacher to preserve the historical cross-modal ranking structure during sequential adaptation.
	The formal stage-wise continual RS-ITR problem is defined in Appendix~A.

	\subsection{Spatial Fusion Adapter}
	\label{subsec:sfa}

	RS images often couple compact targets, such as aircraft, with broad spatial layouts, such as residential areas. Scale changes across continual stages can bias visual adaptation toward newly arrived spatial patterns and weaken previously useful local details or scene-level context. SFA addresses this issue through scale-aware residual adaptation that composes coarse regional cues and fine-grained patch cues while retaining the base image representation.

	Given an RS image $I_i^t$, the visual encoder produces a global token $\mathbf{g}_i^t$, fine patch tokens $\mathbf{P}_i^t$, and a normalized base representation $\mathbf{v}_{i,\mathrm{base}}^t$. SFA pools $\mathbf{P}_i^t$ into coarse regional tokens and performs hierarchical cross-attention in two steps. The coarse tokens first query the fine patch tokens to absorb local evidence; the global token then attends to the concatenated refined coarse and original patch tokens. This hierarchy integrates local objects and regional layouts into a multi-scale global representation.

	An input-dependent gate balances the coarse and fine cues to produce $\mathbf{z}_i^t$, avoiding a fixed weighting across different RS scenes. SFA projects $\mathbf{z}_i^t$ into the shared retrieval space and injects it through a base-preserved residual:
	\begin{equation}
		\mathbf{v}_i^t = \mathrm{Norm}_2\!\left(
		\mathbf{v}_{i,\mathrm{base}}^t
		+\alpha\bigl(\phi_I(\mathbf{z}_i^t)-\mathbf{v}_{i,\mathrm{base}}^t\bigr)
		\right),
	\end{equation}
	where $\phi_I(\cdot)$ is the image-side retrieval projection, $\mathrm{Norm}_2(\cdot)$ denotes $\ell_2$ normalization, and $\alpha$ is a learnable scalar initialized to zero. SFA therefore starts from $\mathbf{v}_{i,\mathrm{base}}^t$ and gradually learns scale-aware corrections, providing spatial adaptation capacity while limiting disturbance to the pretrained cross-modal alignment. Full details are provided in Appendix~B.1.
	
	\subsection{Multi-Expert Semantic Routing}
	\label{subsec:msr}

	Text descriptions may change abruptly across continual stages as new scene categories. 
	Updating a single shared text adapter for all queries can entangle reusable semantics with newly emerging descriptions, causing global text embedding drift. MSR instead decomposes textual adaptation into shared textual semantics and semantically specialized residuals while retaining the base textual representation.

	Given a text query $T_i^t$, the base textual encoder produces the normalized representation $\mathbf{u}_{i,\mathrm{base}}^t$. Each expert applies a lightweight low-rank residual transformation. E1 is an always-active main expert optimized by all queries to accumulate shared textual semantics, whereas E2--E8 are seven auxiliary experts that capture semantically specialized residuals. Using only $\mathbf{u}_{i,\mathrm{base}}^t$ as input, the router selects the Top-2 auxiliary experts and normalizes their routing weights. No task or stage identifier is provided. The routed auxiliary residual is
	\begin{equation}
		\Delta\mathbf{u}_{i,\mathrm{aux}}^t
		= \sum_{m\in\mathcal{K}_i^t}
		\alpha_m(T_i^t)E_m(\mathbf{u}_{i,\mathrm{base}}^t),
	\end{equation}
	where $\mathcal{K}_i^t$ is the selected auxiliary expert set and $E_m(\cdot)$ denotes expert $m$. Sparse routing confines query-dependent adaptation to a small subset of auxiliary paths instead of updating every expert for every description. MSR combines the base textual representation with the main-expert and routed auxiliary residuals:
	\begin{equation}
		\mathbf{u}_i^t = \mathrm{Norm}_2\!\left(
		\mathbf{u}_{i,\mathrm{base}}^t
		+E_{\mathrm{main}}(\mathbf{u}_{i,\mathrm{base}}^t)
		+\Delta\mathbf{u}_{i,\mathrm{aux}}^t\right).
	\end{equation}
	All residual paths are zero-initialized, so the initial output equals the base textual representation. During adaptation, the main expert learns semantics shared across stages, while the routed auxiliary experts absorb more localized semantic changes. This separation increases textual adaptation capacity without requiring task identity and limits abrupt global drift in the shared retrieval space. Low-rank parameterization, routing weights, and initialization details are provided in Appendix~B.2.
	
	\subsection{Bidirectional Ranking Distillation}
	\label{subsec:brd}

	Many existing CL methods inherit classification-oriented objectives and reduce forgetting either through feature-level distillation, which constrains Euclidean feature distances, often with mean squared error, or by distilling classification logits. Such constraints do not directly preserve the information that determines image-text retrieval. RS-ITR depends on the relative ordering of cross-modal similarities: for a given query, its positive candidate must remain ahead of negative candidates. During continual adaptation, individual historical features may remain close to their teacher counterparts while small changes in pairwise similarities reorder positive and negative candidates, thereby reducing retrieval recall. BRD therefore preserves the historical cross-modal ranking structure rather than forcing absolute feature values to remain unchanged.

	After stage $t-1$, BRD freezes the learned model as the teacher $\mathcal{M}_{t-1}$ and retains a compact historical anchor bank $\mathcal{A}_{t-1}$. At stage $t$, both the current model $\mathcal{M}_t$ and the frozen teacher encode a current mini-batch $\mathcal{B}_t$ and a historical anchor batch $\mathcal{B}_a\subset\mathcal{A}_{t-1}$.
	Let $\mathbf{V}_c^t$ and $\mathbf{U}_c^t$ denote the normalized image and text representations of the current batch produced by $\mathcal{M}_t$, and let $\mathbf{V}_a^t$ and $\mathbf{U}_a^t$ denote those of the historical anchor batch. The corresponding teacher representations are denoted by $\mathbf{V}_c^{t-1}$, $\mathbf{U}_c^{t-1}$, $\mathbf{V}_a^{t-1}$, and $\mathbf{U}_a^{t-1}$. For two normalized feature matrices $\mathbf{X}$ and $\mathbf{Y}$, we define the temperature-scaled relation matrix as
	\begin{equation}
		\mathcal{S}(\mathbf{X},\mathbf{Y})
		=
		\frac{\mathbf{X}\mathbf{Y}^{\top}}{\tau},
	\end{equation}
	where $\tau$ is the temperature. Each row of a T2I relation matrix contains one text query's scores against all image candidates, and the ordering of these scores directly determines retrieval recall rather than the feature vector itself. Thus, the current model and frozen teacher may produce different feature vectors for an image without changing its retrieval contribution, provided that its relative rank among image candidates is preserved for every text query. BRD distills this ranking structure through three complementary relations:
	\begin{equation}
	\begin{aligned}
		\mathcal{L}_{\mathrm{T2I}}
		=
		\mathcal{D}_{\mathrm{rel}}
		\left(
		\mathcal{S}(\mathbf{U}_c^t,\mathbf{V}_a^t),
		\mathcal{S}(\mathbf{U}_c^{t-1},\mathbf{V}_a^{t-1})
		\right), \\
		\mathcal{L}_{\mathrm{I2T}}
		=
		\mathcal{D}_{\mathrm{rel}}
		\left(
		\mathcal{S}(\mathbf{V}_c^t,\mathbf{U}_a^t),
		\mathcal{S}(\mathbf{V}_c^{t-1},\mathbf{U}_a^{t-1})
		\right), \\
		\mathcal{L}_{\mathrm{A}}
		=
		\mathcal{D}_{\mathrm{rel}}
		\left(
		\mathcal{S}(\mathbf{V}_a^t,\mathbf{U}_a^t),
		\mathcal{S}(\mathbf{V}_a^{t-1},\mathbf{U}_a^{t-1})
		\right).
	\end{aligned}
	\end{equation}
	Here, $\mathcal{L}_{\mathrm{T2I}}$ preserves the T2I relation between current text queries and historical image anchors, $\mathcal{L}_{\mathrm{I2T}}$ preserves the I2T relation between current images and historical text anchors, and $\mathcal{L}_{\mathrm{A}}$ preserves the anchor-internal relation.
	We instantiate $\mathcal{D}_{\mathrm{rel}}(\cdot,\cdot)$ as mean squared error between corresponding relation matrices. The total BRD regularization is
	\begin{equation}
		\mathcal{L}_{\mathrm{BRD}}
		=
		\lambda_{\mathrm{T2I}}\mathcal{L}_{\mathrm{T2I}}
		+
		\lambda_{\mathrm{I2T}}\mathcal{L}_{\mathrm{I2T}}
		+
		\lambda_{\mathrm{A}}\mathcal{L}_{\mathrm{A}},
	\end{equation}
	where $\lambda_{\mathrm{T2I}}$, $\lambda_{\mathrm{I2T}}$, and $\lambda_{\mathrm{A}}$ balance the three relations. By aligning relation matrices rather than absolute feature values, BRD permits reasonable feature drift while preserving the historical cross-modal ranking structure. Additional BRD details and the overall optimization objective are provided in Appendix~B.3 and Appendix~B.4, respectively.
	
	\section{Experiments}
	\label{sec:experiments}
	
	\begin{table*}[t]
		\centering
		\scriptsize
		\setlength{\tabcolsep}{2.2pt}
		\caption{Comparison with existing methods on RSICD, RSITMD, UCM-Captions, and the semantic-stage RST2I-110K protocol. RST2I-110K uses unified all-seen galleries. Higher mR, C-mR, and F-mR are better; a negative F-mR indicates degradation. ``--'' denotes unavailable or undefined entries.}
		\label{tab:sota_comparison}
		\resizebox{\textwidth}{!}{%
			\begin{tabular}{@{}l!{\vrule width 0.5pt}ccc!{\vrule width 0.5pt}*{12}{c}@{}}
				\toprule
				\multicolumn{1}{c!{\vrule width 0.5pt}}{\multirow{3}{*}{Method}}
				& \multirow{2}{*}{RSICD}
				& \multirow{2}{*}{RSITMD}
				& \multirow{2}{*}{UCM}
				& \multicolumn{12}{c}{RST2I-110K} \\
				\cmidrule(l){5-16}
				& & &
				& \multicolumn{2}{c}{Task 1}
				& \multicolumn{2}{c}{Task 2}
				& \multicolumn{2}{c}{Task 3}
				& \multicolumn{2}{c}{Task 4}
				& \multicolumn{2}{c}{Task 5}
				& \multicolumn{2}{c}{Task 6} \\
				\cmidrule(lr){5-6}\cmidrule(lr){7-8}\cmidrule(lr){9-10}
				\cmidrule(lr){11-12}\cmidrule(lr){13-14}\cmidrule(l){15-16}
				& mR & mR & mR
				& C-mR & F-mR & C-mR & F-mR & C-mR & F-mR
				& C-mR & F-mR & C-mR & F-mR & C-mR & F-mR \\
				\midrule
				\multicolumn{16}{c}{\textit{Remote sensing methods}} \\
				\midrule
				GaLR {\color{gray}\scriptsize\itshape TGRS'22}
				& 18.24 & 30.91 & 48.21
				& 3.73 & --
				& 6.99 & \textcolor{red}{$-2.45$}
				& 14.96 & \underline{\textcolor{red}{$-0.74$}}
				& 16.05 & \textcolor{red}{$-1.01$}
				& 10.91 & \textcolor{red}{$-3.71$}
				& 23.04 & \textcolor{red}{$-1.06$} \\
				PIR {\color{gray}\scriptsize\itshape ACM MM'23}
				& 24.40 & 38.76 & 52.73
				& 4.12 & --
				& 7.83 & \textcolor{red}{$-2.04$}
				& 18.34 & \textcolor{red}{$-0.93$}
				& 20.66 & \underline{\textcolor{red}{$-0.15$}}
				& 14.78 & \textcolor{red}{$-2.27$}
				& 29.29 & \underline{\textcolor{red}{$-0.12$}} \\
				CFTIR {\color{gray}\scriptsize\itshape GRSL'25}
				& 37.13 & 43.87 & 52.10
				& 4.04 & --
				& 7.91 & \textcolor{red}{$-2.46$}
				& 17.49 & \textcolor{red}{$-0.81$}
				& 21.04 & \textcolor{red}{$-0.64$}
				& 15.31 & \underline{\textcolor{red}{$-1.89$}}
				& 27.58 & \textcolor{red}{$-1.43$} \\
				\midrule
				\multicolumn{16}{c}{\textit{CLIP-based methods}} \\
				\midrule
				CLIP-ft {\color{gray}\scriptsize\itshape ICML'21}
				& 35.82 & 48.56 & 52.60
				& 12.89 & --
				& 18.89 & \textcolor{red}{$-5.00$}
				& 34.98 & \textcolor{red}{$-2.45$}
				& 34.52 & \textcolor{red}{$-2.50$}
				& 27.07 & \textcolor{red}{$-4.85$}
				& 43.23 & \textcolor{red}{$-3.60$} \\
				CoOp {\color{gray}\scriptsize\itshape IJCV'22}
				& 20.51 & 27.85 & 34.98
				& 6.07 & --
				& 10.75 & \textcolor{red}{$-2.21$}
				& 19.71 & \textcolor{red}{$-1.67$}
				& 21.17 & \textcolor{red}{$-1.41$}
				& 13.35 & \textcolor{red}{$-3.28$}
				& 26.93 & \textcolor{red}{$-2.17$} \\
				VPT {\color{gray}\scriptsize\itshape ECCV'22}
				& 25.95 & 34.14 & 35.64
				& 6.00 & --
				& 9.34 & \textcolor{red}{$-2.44$}
				& 18.88 & \textcolor{red}{$-1.71$}
				& 19.11 & \textcolor{red}{$-1.38$}
				& 14.01 & \textcolor{red}{$-2.46$}
				& 26.29 & \textcolor{red}{$-1.53$} \\
				MaPLe {\color{gray}\scriptsize\itshape CVPR'23}
				& 29.31 & 41.23 & 38.78
				& 8.50 & --
				& 13.90 & \textcolor{red}{$-3.89$}
				& 24.94 & \textcolor{red}{$-2.65$}
				& 25.92 & \textcolor{red}{$-2.61$}
				& 18.54 & \textcolor{red}{$-3.83$}
				& 32.87 & \textcolor{red}{$-2.83$} \\
				RemoteCLIP {\color{gray}\scriptsize\itshape TGRS'24}
				& 36.35 & 50.52 & \textbf{56.36}
				& \textbf{15.33} & --
				& \underline{20.14} & \underline{\textcolor{red}{$-0.97$}}
				& 33.57 & \textcolor{red}{$-1.29$}
				& \underline{36.48} & \textcolor{red}{$-1.43$}
				& \underline{29.44} & \textcolor{red}{$-2.73$}
				& 44.59 & \textcolor{red}{$-4.22$} \\
				AIR {\color{gray}\scriptsize\itshape ACM MM'24}
				& 36.24 & 50.22 & 48.09
				& \underline{13.12} & --
				& 18.97 & \textcolor{red}{$-4.94$}
				& \underline{35.11} & \textcolor{red}{$-2.37$}
				& 35.41 & \textcolor{red}{$-2.27$}
				& 28.25 & \textcolor{red}{$-3.84$}
				& \underline{45.44} & \textcolor{red}{$-2.87$} \\
				FSSN {\color{gray}\scriptsize\itshape TGRS'25}
				& 37.39 & 51.28 & 46.80
				& 11.59 & --
				& 16.61 & \textcolor{red}{$-5.48$}
				& 29.90 & \textcolor{red}{$-2.56$}
				& 29.80 & \textcolor{red}{$-2.05$}
				& 22.84 & \textcolor{red}{$-5.62$}
				& 37.08 & \textcolor{red}{$-3.39$} \\
				TriSim {\color{gray}\scriptsize\itshape CVPR'26}
				& \underline{37.55} & \textbf{51.35} & 49.31
				& 11.58 & --
				& 16.06 & \textcolor{red}{$-4.88$}
				& 29.56 & \textcolor{red}{$-2.63$}
				& 30.05 & \textcolor{red}{$-1.71$}
				& 22.71 & \textcolor{red}{$-5.11$}
				& 38.86 & \textcolor{red}{$-2.86$} \\
				\midrule
				\textbf{DARAD \color{gray}\scriptsize\itshape (ours)}
				& \textbf{37.92} & \underline{51.31} & \underline{53.28}
				& 12.16 & --
				& \textbf{21.17} & \textcolor{green!50!black}{$\mathbf{+0.44}$}
				& \textbf{39.58} & \textcolor{green!50!black}{$\mathbf{+0.76}$}
				& \textbf{39.06} & \textcolor{green!50!black}{$\mathbf{+0.24}$}
				& \textbf{30.67} & \textcolor{green!50!black}{$\mathbf{+0.39}$}
				& \textbf{50.73} & \textcolor{green!50!black}{$\mathbf{+1.01}$} \\
				\bottomrule
			\end{tabular}%
		}
	\end{table*}
	
	\subsection{Experimental Setup}
	\label{subsec:experimental_setup}

	\noindent\textbf{Datasets and protocol.}
	We evaluate DARAD on RST2I-110K~\cite{zhang2026any2rsi} and three manually annotated RS-ITR benchmarks, including RSICD~\cite{lu2018rsicd}, RSITMD~\cite{yuan2022amfmn}, and UCM-Captions~\cite{yang2010ucmerced,qu2016ucm}. RST2I-110K serves as the main continual RS-ITR benchmark due to its large scale and broad semantic coverage, and is organized into six sequential semantic stages. At stage $t$, the model learns only the newly arrived data. 
	Detailed statistics and stage splits are provided in Appendix~\appDatasetProtocol{}.

	\noindent\textbf{Evaluation metrics.}
	We use mean Recall (mR), averaging Recall@1, 5, and 10 over image-to-text (I2T) and text-to-image (T2I) retrieval. Let $\mathcal{Q}_k$ be the queries originating from stage $k$ and $\mathcal{G}_{\leq t}=\bigcup_{j=1}^{t}\mathcal{G}_j$ the unified all-seen gallery at stage $t$ (the corresponding image or text candidates for T2I or I2T). We define $a_{t,k}=\operatorname{mR}(\mathcal{M}_t,\mathcal{Q}_k,\mathcal{G}_{\leq t})$, where $1\leq k\leq t\leq T$ and $T=6$. Current mR (C-mR) at stage $t$ is $a_{t,t}$, Avg. C-mR is $\frac{1}{T}\sum_{t=1}^{T}a_{t,t}$, and Final mR is $\frac{1}{T}\sum_{k=1}^{T}a_{T,k}$. F-mR measures the signed historical performance change under the all-seen gallery:
	\begin{equation}
		\mathrm{F\text{-}mR}_{t}
		=
		\frac{1}{t-1}
		\sum_{k=1}^{t-1}
		\left(a_{t,k}-a_{k,k}\right),
		\qquad t\geq 2.
	\end{equation}
	
	\noindent\textbf{Implementation details.}
	We use a ViT-B/32 dual-encoder retrieval model and resize all images to $224\times224$ with CLIP normalization. For SFA, the coarse spatial grid is $3\times3$ with 8-head cross-attention. MSR uses one always-active main expert and seven auxiliary experts (eight in total), with Top-2 routing over E2--E8 and a low-rank dimension of 16. For BRD, we retain 1,000 historical anchors per stage by herding. The model is optimized with AdamW using learning rates of $3\times10^{-6}$ and $10^{-5}$ for the base model and adaptation modules, respectively, with a batch size of 64. 
	\begin{table*}[t]
	\centering
	\scriptsize
	\setlength{\tabcolsep}{2.1pt}
	\renewcommand{\arraystretch}{1.18}
	\setlength{\heavyrulewidth}{0.9pt}
	\setlength{\lightrulewidth}{0.45pt}
	\setlength{\cmidrulewidth}{0.4pt}
	\caption{Comparison with existing CL methods on the six semantic stages of RST2I-110K using unified all-seen galleries. Mean results over five runs under fixed O1 are reported, with standard deviations as subscripts and rankings based on means. Task 1 reports only C-mR because F-mR requires historical tasks.}
	\label{tab:continual_comparison}
	\resizebox{\textwidth}{!}{%
		\begin{tabular}{@{}l!{\vrule width 0.65pt}*{11}{c}@{}}
			\toprule
			\multicolumn{1}{c!{\vrule width 0.65pt}}{\multirow{3}{*}{Method}}
			& \multicolumn{11}{c}{RST2I-110K} \\
			\cmidrule(l){2-12}
			& \multicolumn{1}{c}{Task 1}
			& \multicolumn{2}{c}{Task 2}
			& \multicolumn{2}{c}{Task 3}
			& \multicolumn{2}{c}{Task 4}
			& \multicolumn{2}{c}{Task 5}
			& \multicolumn{2}{c}{Task 6} \\
			\cmidrule(lr){2-2}\cmidrule(lr){3-4}\cmidrule(lr){5-6}
			\cmidrule(lr){7-8}\cmidrule(lr){9-10}\cmidrule(l){11-12}
			& C-mR & C-mR & F-mR & C-mR & F-mR
			& C-mR & F-mR & C-mR & F-mR & C-mR & F-mR \\
			\midrule
			SeqFT {\color{gray}\scriptsize\itshape Baseline}
			& $\mathbf{15.33}_{\scriptscriptstyle\pm0.24}$
			& $\underline{20.14}_{\scriptscriptstyle\pm0.31}$ & \textcolor{red}{$\underline{-0.97}_{\scriptscriptstyle\pm0.19}$}
			& $\underline{33.57}_{\scriptscriptstyle\pm0.29}$ & \textcolor{red}{$-1.29_{\scriptscriptstyle\pm0.23}$}
			& $\underline{36.48}_{\scriptscriptstyle\pm0.34}$ & \textcolor{red}{$-1.43_{\scriptscriptstyle\pm0.26}$}
			& $\underline{29.44}_{\scriptscriptstyle\pm0.37}$ & \textcolor{red}{$-2.73_{\scriptscriptstyle\pm0.32}$}
			& $\underline{44.59}_{\scriptscriptstyle\pm0.42}$ & \textcolor{red}{$-4.22_{\scriptscriptstyle\pm0.39}$} \\
			SeqLoRA {\color{gray}\scriptsize\itshape ICLR'22}
			& $11.16_{\scriptscriptstyle\pm0.21}$
			& $16.93_{\scriptscriptstyle\pm0.27}$ & \textcolor{red}{$-4.59_{\scriptscriptstyle\pm0.34}$}
			& $31.83_{\scriptscriptstyle\pm0.33}$ & \textcolor{red}{$-2.18_{\scriptscriptstyle\pm0.25}$}
			& $32.71_{\scriptscriptstyle\pm0.31}$ & \textcolor{red}{$-2.43_{\scriptscriptstyle\pm0.28}$}
			& $24.93_{\scriptscriptstyle\pm0.36}$ & \textcolor{red}{$-4.01_{\scriptscriptstyle\pm0.37}$}
			& $41.72_{\scriptscriptstyle\pm0.40}$ & \textcolor{red}{$-2.59_{\scriptscriptstyle\pm0.30}$} \\
			C-CLIP {\color{gray}\scriptsize\itshape ICLR'25}
			& $10.19_{\scriptscriptstyle\pm0.18}$
			& $7.90_{\scriptscriptstyle\pm0.23}$ & \textcolor{red}{$-1.80_{\scriptscriptstyle\pm0.20}$}
			& $19.06_{\scriptscriptstyle\pm0.28}$ & \textcolor{red}{$-0.46_{\scriptscriptstyle\pm0.13}$}
			& $22.83_{\scriptscriptstyle\pm0.30}$ & \textcolor{green!50!black}{$\mathbf{+0.71}_{\scriptscriptstyle\pm0.12}$}
			& $15.94_{\scriptscriptstyle\pm0.27}$ & \textcolor{green!50!black}{$\underline{+0.58}_{\scriptscriptstyle\pm0.11}$}
			& $34.23_{\scriptscriptstyle\pm0.35}$ & \textcolor{green!50!black}{$+0.74_{\scriptscriptstyle\pm0.15}$} \\
			GIFT {\color{gray}\scriptsize\itshape CVPR'25}
			& $10.18_{\scriptscriptstyle\pm0.13}$
			& $7.64_{\scriptscriptstyle\pm0.24}$ & \textcolor{red}{$-1.31_{\scriptscriptstyle\pm0.17}$}
			& $18.87_{\scriptscriptstyle\pm0.26}$ & \textcolor{green!50!black}{$\underline{+0.32}_{\scriptscriptstyle\pm0.11}$}
			& $22.18_{\scriptscriptstyle\pm0.29}$ & \textcolor{red}{$-0.19_{\scriptscriptstyle\pm0.14}$}
			& $14.58_{\scriptscriptstyle\pm0.32}$ & \textcolor{green!50!black}{$\mathbf{+0.62}_{\scriptscriptstyle\pm0.10}$}
			& $33.12_{\scriptscriptstyle\pm0.34}$ & \textcolor{green!50!black}{$\underline{+0.93}_{\scriptscriptstyle\pm0.13}$} \\
			MG-CLIP {\color{gray}\scriptsize\itshape ICCV'25}
			& $10.07_{\scriptscriptstyle\pm0.20}$
			& $7.83_{\scriptscriptstyle\pm0.25}$ & \textcolor{red}{$-1.78_{\scriptscriptstyle\pm0.19}$}
			& $19.10_{\scriptscriptstyle\pm0.30}$ & \textcolor{red}{$-0.57_{\scriptscriptstyle\pm0.14}$}
			& $22.93_{\scriptscriptstyle\pm0.28}$ & \textcolor{green!50!black}{$\underline{+0.50}_{\scriptscriptstyle\pm0.12}$}
			& $16.14_{\scriptscriptstyle\pm0.26}$ & \textcolor{green!50!black}{$+0.34_{\scriptscriptstyle\pm0.11}$}
			& $34.00_{\scriptscriptstyle\pm0.37}$ & \textcolor{green!50!black}{$+0.72_{\scriptscriptstyle\pm0.14}$} \\
			ConDU {\color{gray}\scriptsize\itshape ICLR'26}
			& $10.24_{\scriptscriptstyle\pm0.14}$
			& $10.59_{\scriptscriptstyle\pm0.27}$ & \textcolor{red}{$-1.91_{\scriptscriptstyle\pm0.21}$}
			& $24.66_{\scriptscriptstyle\pm0.31}$ & \textcolor{green!50!black}{$+0.29_{\scriptscriptstyle\pm0.13}$}
			& $27.86_{\scriptscriptstyle\pm0.29}$ & \textcolor{red}{$-0.11_{\scriptscriptstyle\pm0.15}$}
			& $20.70_{\scriptscriptstyle\pm0.33}$ & \textcolor{green!50!black}{$+0.41_{\scriptscriptstyle\pm0.12}$}
			& $39.46_{\scriptscriptstyle\pm0.36}$ & \textcolor{green!50!black}{$+0.50_{\scriptscriptstyle\pm0.14}$} \\
			\specialrule{0.7pt}{1.5pt}{1.5pt}
			\textbf{DARAD} {\color{gray}\scriptsize\itshape (ours)}
			& $\underline{12.16}_{\scriptscriptstyle\pm0.19}$
			& $\mathbf{21.17}_{\scriptscriptstyle\pm0.20}$ & \textcolor{green!50!black}{$\mathbf{+0.44}_{\scriptscriptstyle\pm0.08}$}
			& $\mathbf{39.58}_{\scriptscriptstyle\pm0.26}$ & \textcolor{green!50!black}{$\mathbf{+0.76}_{\scriptscriptstyle\pm0.10}$}
			& $\mathbf{39.06}_{\scriptscriptstyle\pm0.24}$ & \textcolor{green!50!black}{$+0.24_{\scriptscriptstyle\pm0.09}$}
			& $\mathbf{30.67}_{\scriptscriptstyle\pm0.28}$ & \textcolor{green!50!black}{$+0.39_{\scriptscriptstyle\pm0.10}$}
			& $\mathbf{50.73}_{\scriptscriptstyle\pm0.30}$ & \textcolor{green!50!black}{$\mathbf{+1.01}_{\scriptscriptstyle\pm0.12}$} \\
			\bottomrule
		\end{tabular}%
	}
\end{table*}
	
	\subsection{Comparison with State-of-the-Art Methods}
	\label{subsec:sota_comparison}

	We compare DARAD with two groups of baselines. Table~\ref{tab:sota_comparison} includes RS-specific retrieval methods (GaLR, PIR, and CFTIR)~\cite{yuan2022galr,yuan2023pir,zhou2024cftir}, together with CLIP-based adaptation and recent RS-ITR methods (CLIP-ft, CoOp, VPT, MaPLe, RemoteCLIP, AIR, FSSN, and TriSim)~\cite{radford2021clip,zhou2022coop,jia2022vpt,khattak2023maple,liu2024remoteclip,cheng2024air,zheng2025fssn,zheng2026trisim}. Table~\ref{tab:continual_comparison} evaluates continual learning on RST2I-110K using SeqFT, parameter-efficient SeqLoRA, C-CLIP, which directly considers continual image-text retrieval, and the VLM-CL frameworks GIFT, MG-CLIP, and ConDU adapted to the same retrieval protocol~\cite{liu2025cclip,wu2025gift,huang2025mgclip,gao2026condu}. Baseline configurations and retrieval adaptations are detailed in Appendix~\appControlledAblations{}.
		\begin{table*}[t]
	\centering
	\scriptsize
	\setlength{\tabcolsep}{3.0pt}
	\renewcommand{\arraystretch}{1.18}
	\setlength{\heavyrulewidth}{0.9pt}
	\setlength{\lightrulewidth}{0.45pt}
	\setlength{\cmidrulewidth}{0.4pt}
	\caption{Main component ablation of DARAD on the six semantic stages of RST2I-110K using unified all-seen galleries. Each stage reports C-mR and signed F-mR; F-mR is undefined for Task 1.}
	\label{tab:main_ablation}
		\begin{tabular*}{\textwidth}{@{\extracolsep{\fill}}l!{\vrule width 0.65pt}*{12}{c}@{}}
			\toprule
			\multicolumn{1}{c!{\vrule width 0.65pt}}{\multirow{3}{*}{Variant}}
			& \multicolumn{12}{c}{RST2I-110K} \\
			\cmidrule(l){2-13}
			& \multicolumn{2}{c}{Task 1}
			& \multicolumn{2}{c}{Task 2}
			& \multicolumn{2}{c}{Task 3}
			& \multicolumn{2}{c}{Task 4}
			& \multicolumn{2}{c}{Task 5}
			& \multicolumn{2}{c}{Task 6} \\
			\cmidrule(lr){2-3}\cmidrule(lr){4-5}\cmidrule(lr){6-7}
			\cmidrule(lr){8-9}\cmidrule(lr){10-11}\cmidrule(l){12-13}
			& C-mR & F-mR & C-mR & F-mR & C-mR & F-mR
			& C-mR & F-mR & C-mR & F-mR & C-mR & F-mR \\
			\midrule
			RemoteCLIP
			& 15.33 & -- & 20.14 & \textcolor{red}{$-0.97$}
			& 33.57 & \textcolor{red}{$-1.29$}
			& 36.48 & \textcolor{red}{$-1.43$}
			& 29.44 & \textcolor{red}{$-2.73$}
			& 44.59 & \textcolor{red}{$-4.22$} \\
			w/o SFA
			& 11.84 & -- & 20.42 & \textcolor{green!50!black}{$+0.28$}
			& 37.21 & \textcolor{green!50!black}{$+0.47$}
			& 37.02 & \textcolor{green!50!black}{$+0.13$}
			& 29.12 & \textcolor{green!50!black}{$+0.21$}
			& 48.36 & \textcolor{green!50!black}{$+0.68$} \\
			w/o MSR
			& 11.67 & -- & 19.86 & \textcolor{green!50!black}{$+0.19$}
			& 36.84 & \textcolor{green!50!black}{$+0.32$}
			& 36.55 & \textcolor{green!50!black}{$+0.06$}
			& 28.74 & \textcolor{green!50!black}{$+0.15$}
			& 47.92 & \textcolor{green!50!black}{$+0.51$} \\
			w/o BRD
			& 12.16 & -- & 21.04 & \textcolor{red}{$-0.42$}
			& 38.91 & \textcolor{red}{$-0.58$}
			& 38.52 & \textcolor{red}{$-0.81$}
			& 30.18 & \textcolor{red}{$-1.14$}
			& 49.86 & \textcolor{red}{$-1.62$} \\
			\specialrule{0.7pt}{1.5pt}{1.5pt}
			\textbf{DARAD}
			& 12.16 & -- & 21.17 & \textcolor{green!50!black}{$+0.44$}
			& 39.58 & \textcolor{green!50!black}{$+0.76$}
			& 39.06 & \textcolor{green!50!black}{$+0.24$}
			& 30.67 & \textcolor{green!50!black}{$+0.39$}
			& 50.73 & \textcolor{green!50!black}{$+1.01$} \\
			\bottomrule
		\end{tabular*}
\end{table*}

	Table~\ref{tab:sota_comparison} reports standard retrieval performance and stage-wise continual retrieval results. On the three classical benchmarks, DARAD achieves the best mR on RSICD and the second-best mR on RSITMD and UCM-Captions, indicating that the proposed continual adaptation design remains competitive under standard static retrieval protocols. On RST2I-110K, several strong CLIP-based methods achieve competitive C-mR on individual stages, but their F-mR values are consistently negative after new semantic stages arrive. In contrast, DARAD obtains the best C-mR from Task 2 to Task 6 and maintains positive F-mR values from $+0.24$ to $+1.01$. This suggests that DARAD improves adaptation to newly arrived RS image-text pairs while better preserving historical-query retrieval performance.

	Table~\ref{tab:continual_comparison} further isolates CL behavior on RST2I-110K. SeqFT preserves strong plasticity in early stages, but its F-mR remains negative after every subsequent stage, showing declining historical retrieval as the unified gallery expands. SeqLoRA provides parameter-efficient adaptation but still suffers from historical degradation. Recent VLM CL methods, including C-CLIP, GIFT, MG-CLIP, and ConDU, improve historical retention in several later stages.
	
	DARAD achieves the highest C-mR on Tasks 2--6 while keeping positive F-mR throughout Tasks 2--6, indicating a more favorable balance between new-stage adaptation and historical retrieval performance retention. 
	Appendix~\appOrderRobustness{} further evaluates two additional fixed stage permutations. Across the three orders, DARAD obtains $32.11\pm0.13$ Avg. C-mR, $+0.53\pm0.04$ Avg. F-mR, and $0.62\pm0.06$ maximum forgetting, indicating limited sensitivity to the evaluated semantic-stage order. Appendix~\appCrossDataset{} additionally evaluates a cross-dataset stream, where DARAD obtains 47.46 mR and near-zero Avg. F-mR ($+0.01$) under dataset-level distribution changes.

	\subsection{Ablation Studies}
	\label{subsec:ablation}

	We conduct ablation studies on RST2I-110K to examine the contribution of each component in DARAD. Table~\ref{tab:main_ablation} reports removal-based component ablations over the six semantic stages. All variants use the same stage order, training schedule, and evaluation protocol. Controlled comparisons and mechanism-level ablations are provided in Appendix~\appControlledAblations{}.

	Table~\ref{tab:main_ablation} shows that each component contributes to continual RS-ITR from a different perspective. Removing BRD keeps relatively high C-mR but changes all F-mR values to negative values, confirming that BRD is critical for preserving the historical ranking structure. Removing SFA or MSR still keeps positive F-mR because BRD is retained, but both variants consistently reduce C-mR compared with full DARAD, indicating that scale-aware visual adaptation and semantically routed adaptation are important for learning newly arrived semantics. 
	Mechanism-level ablations in Appendix~\appControlledAblations{} provide further evidence beyond component removal: coarse--fine spatial fusion improves both local-object and global-layout retrieval while reducing visual drift, and sparse semantic routing reduces historical text drift compared with main-only and dense-routing variants. 
	Across five matched seeds under the same anchor and teacher budgets, BRD improves Avg. C-mR and Avg. F-mR by 0.25 and 0.26 points, respectively, over replay combined with feature-level distillation; the Avg. F-mR retention gain is statistically significant (two-sided paired $t$-test, $p=0.0013$). With fixed total budgets of 1,000 and 2,500 anchors, BRD maintains positive Avg. F-mR ($+0.09$ and $+0.25$) and remains above the matched replay baseline in both aggregate metrics. The complete memory-matched, fixed-budget, and significance analyses are reported in Appendix~\appControlledAblations{}.
	
		\begin{table*}[t]
	\centering
	\small
	\setlength{\tabcolsep}{4pt}
	\renewcommand{\arraystretch}{1.12}
	\caption{Stage-wise historical ranking preservation on RST2I-110K under the default stage order (O1). At each transition $t=2,\ldots,6$, the previous-stage and updated models rank the same pre-update historical gallery.}
	\label{tab:ranking_consistency}
	\begin{tabular*}{\textwidth}{@{\extracolsep{\fill}}l!{\vrule width 0.5pt}cccccc@{}}
		\toprule
		\multicolumn{1}{c!{\vrule width 0.5pt}}{\multirow{3}{*}{Retention mechanism}}
		& \multicolumn{6}{c}{RST2I-110K} \\
		\cmidrule(l){2-7}
		
		& \multicolumn{3}{c}{T2I}
		& \multicolumn{3}{c}{I2T} \\
		\cmidrule(lr){2-4}\cmidrule(lr){5-7}
		& Kendall $\tau$ & Spearman $\rho$ & Overlap@10
		& Kendall $\tau$ & Spearman $\rho$ & Overlap@10 \\
		\midrule
		w/o BRD
		& 0.486 & 0.662 & 0.614
		& 0.514 & 0.684 & 0.642 \\
		Feature KD
		& 0.591 & 0.728 & 0.722
		& 0.617 & 0.771 & 0.731 \\
		Replay + Feature KD
		& \underline{0.647} & \underline{0.791} & \underline{0.751}
		& \underline{0.649} & \underline{0.803} & \underline{0.764} \\
		\textbf{BRD (DARAD)}
		& \textbf{0.724} & \textbf{0.857} & \textbf{0.813}
		& \textbf{0.779} & \textbf{0.901} & \textbf{0.862} \\
		\bottomrule
	\end{tabular*}
\end{table*}
	\subsection{Mechanism Analysis and Visualization}
	\label{subsec:mechanism_visualization}




	\subsubsection{Spatial Response of SFA}
	\label{subsubsec:sfa_spatial_visualization}

\begin{figure}[t]
	\centering
	\includegraphics[width=\linewidth]{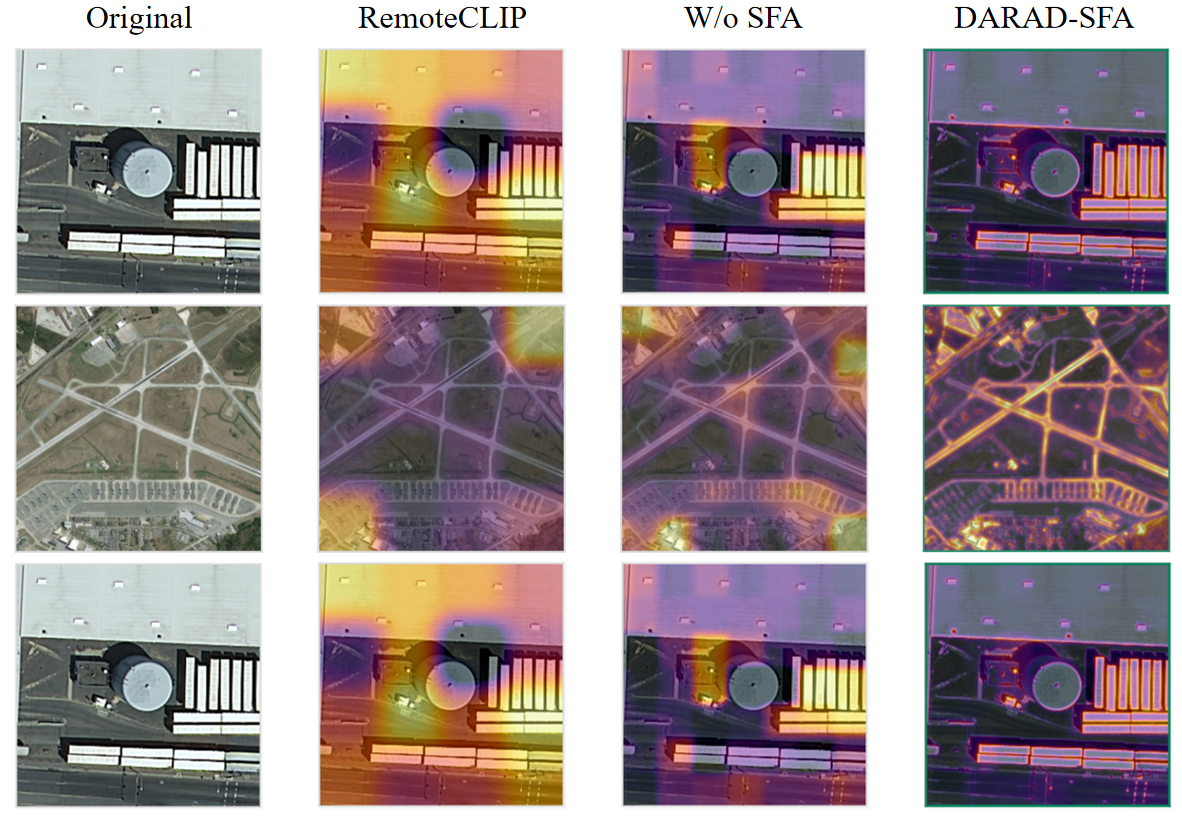}
	\caption{Spatial response comparison of the original images, RemoteCLIP, DARAD without SFA, and DARAD-SFA on representative RS scenes.}
	\label{fig:sfa_spatial_response}
\end{figure}

	Fig.~\ref{fig:sfa_spatial_response} compares spatial response maps on representative RS scenes to illustrate the effect of SFA. RemoteCLIP tends to produce broad region-level responses, and the w/o SFA variant captures part of the relevant structure but remains less precise around small objects and local boundaries. In contrast, DARAD-SFA produces more concentrated responses over retrieval-relevant structures, such as aircraft and nearby vehicles, storage tanks and surrounding containers, and sports-field boundaries. 

\begin{figure}[t]
	\centering
	\includegraphics[width=\linewidth]{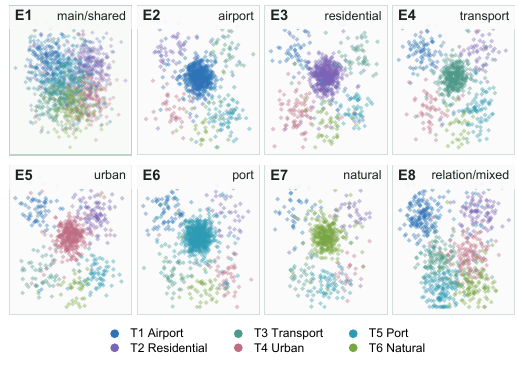}
	\caption{Final-stage MSR routing. E1 is the always-active expert and E2--E8 are sparsely routed auxiliary experts.}
	\label{fig:msr_final_expert_routing}
\end{figure}
	
	\subsubsection{Final-stage Expert Routing of MSR}
	\label{subsubsec:msr_final_routing}

	Fig.~\ref{fig:msr_final_expert_routing} visualizes final-stage MSR routing on the seen RST2I-110K test split. E1 is the always-active shared expert and E2--E8 visualize auxiliary routing responses. 
	E1 and E8 contain mixed colors, suggesting that they capture shared and relation-oriented semantics across stages.
	This indicates that MSR does not absorb newly arrived text semantics through a single global update, but routes semantically related residuals to specialized experts. The complete routing evolution over continual stages is provided in Appendix~\appVisualAnalysis{}.
	
	\subsubsection{Historical Ranking Consistency of BRD}
	\label{subsubsec:brd_ranking_consistency}

	To directly assess historical ranking preservation, we compare the frozen previous-stage model $\mathcal{M}_{t-1}$ with the updated model $\mathcal{M}_t$ at each transition $t=2,\ldots,6$. Kendall's $\tau$ measures pairwise ordering agreement, Spearman's $\rho$ measures rank displacement, and Overlap@10 measures how many previous-model Top-10 candidates remain in the updated-model Top-10. To emphasize retrieval-relevant candidates rather than the long gallery tail, $\tau$ and $\rho$ are computed over the union of the two models' Top-100 candidates. Formal definitions are provided in Appendix~\appControlledAblations{}.

	Table~\ref{tab:ranking_consistency} shows that feature-level distillation improves ranking consistency over the unconstrained variant, and historical replay provides a further gain. BRD achieves the highest consistency in both retrieval directions. Compared with Replay + Feature KD, BRD improves Kendall's $\tau$, Spearman's $\rho$, and Overlap@10 by 0.077, 0.066, and 0.062 in T2I, and by 0.130, 0.098, and 0.098 in I2T, respectively. These results indicate that directly aligning cross-modal relation matrices preserves historical retrieval order more effectively than point-wise feature matching and replay.

	\section{Discussion and Conclusion}
\label{sec:discussion_conclusion}

This work studies semantic-stage continual RS-ITR, where sequential adaptation may disrupt historical visual-textual representations and cross-modal rankings.
DARAD combines two complementary adapters with ranking-aware distillation: SFA supports scale-aware visual adaptation, MSR accommodates evolving textual semantics through sparse expert routing, and BRD constrains changes in historical bidirectional ranking structures. Experiments on three standard RS-ITR benchmarks show that DARAD remains competitive under static retrieval settings. Under the controlled six-stage RST2I-110K protocol, DARAD achieves a favorable balance between current-stage retrieval and historical retention across the evaluated task orders.
\bibliography{references}

\end{document}